\pdfoutput=1
\documentclass[11pt]{article}
\usepackage{authblk} % for fancy author format

\usepackage{EMNLP2022}

\usepackage{times}
\usepackage{latexsym}

\usepackage[T1]{fontenc}
\usepackage[utf8]{inputenc}

\usepackage{microtype}

\usepackage{inconsolata}

\usepackage{float} % for table / figure placement
\usepackage{booktabs} % for nicer table format
\usepackage{graphicx} % for managing images
\usepackage{bm} % for bold maths
\usepackage{mdframed} %for framing
\usepackage{dblfloatfix} % for enabling figures at the bottom of page

\newenvironment{RecBox}[1]
  {\mdfsetup{
    frametitle={\colorbox{white}{\space#1\space}},
    innerleftmargin = 0.25cm, innerrightmargin = 0.25cm, innertopmargin = 0cm, innerbottommargin = 0.25cm,
    frametitleaboveskip=-\ht\strutbox,
    frametitlealignment={\hspace{-5pt}},
    skipabove=10pt,
    skipbelow=7pt
    }
  \begin{mdframed}%
  }
  {\end{mdframed}}

\title{Data-Efficient Strategies for Expanding Hate Speech Detection into Under-Resourced Languages}

\author[1]{\textbf{Paul R\"ottger}}
\author[2]{\textbf{Debora Nozza}}
\author[3]{\textbf{Federico Bianchi}}
\author[2]{\textbf{Dirk Hovy}}
\affil[1]{University of Oxford}
\affil[2]{Bocconi University}
\affil[3]{Stanford University}

\begin{document}
\maketitle
\begin{abstract}
Hate speech is a global phenomenon, but most hate speech datasets so far focus on English-language content.
This hinders the development of more effective hate speech detection models in hundreds of languages spoken by billions across the world.
More data is needed, but annotating hateful content is expensive, time-consuming and potentially harmful to annotators.
To mitigate these issues, we explore data-efficient strategies for expanding hate speech detection into under-resourced languages.
In a series of experiments with mono- and multilingual models across five non-English languages, we find that
1) a small amount of target-language fine-tuning data is needed to achieve strong performance,
2) the benefits of using more such data decrease exponentially, and
3)~initial fine-tuning on readily-available English data can partially substitute target-language data and improve model generalisability.
Based on these findings, we formulate actionable recommendations for hate speech detection in low-resource language settings.

\textcolor{red}{\textbf{Content warning}:
This article contains illustrative examples of hateful language.}
\end{abstract}

%%%%%%%%%%%%%%%%%%%%%%%%%%%%%%%%%%%%%%%%%%%%%%%%%%%%%%%%%%%%%%%%%%%%%%%%%%%%%%%%%%%%
%%%%%%%%%%%%%%%%%%%%%%%%%%%%%%%%%%%%%%%%%%%%%%%%%%%%%%%%%%%%%%%%%%%%%%%%%%%%%%%%%%%%
\section{Introduction} \label{sec: intro}

%Hate speech detection models play a key role in online content moderation and also enable scientific analysis and monitoring of online hate.
Hate speech is a global phenomenon, but most hate speech datasets so far focus on English-language content \citep{vidgen2020directions, poletto2021resources}.
This hinders the development of effective models for detecting hate speech in other languages.
As a consequence, billions of non-English speakers across the world are less protected against online hate, and even giant social media platforms have clear language gaps in their content moderation systems \citep{simonite2021facebook, marinescu2021facebook}.

\begin{figure}[t]
\centering
\includegraphics[width=0.35\textwidth]{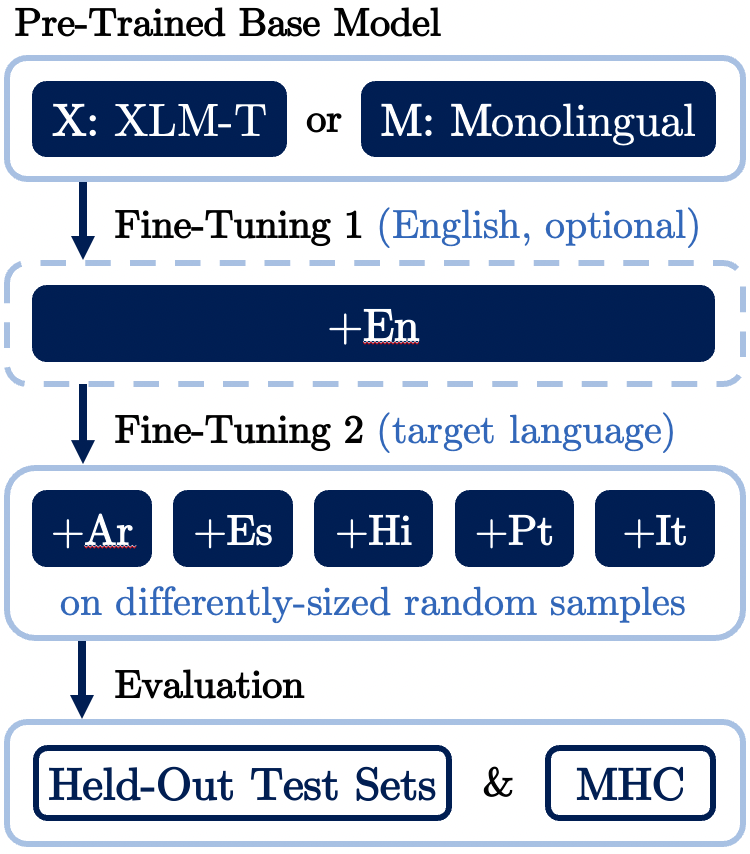}
\caption{Overview of our experimental setup.
We use ISO 639-1 codes to denote the different languages.
MHC is Multilingual HateCheck \citep{rottger2022mhc}.
}
\label{fig: overview}
\end{figure}

Zero-shot cross-lingual transfer, where large multilingual language models are fine-tuned on one source language and then applied to another target language, may appear like a potential solution to the issue of language-specific resource scarcity.
However, while this approach performs well on some tasks \citep{conneau2020xlmr, barbieri2022xlmt}, it fails on many others \citep{lauscher2020zero, hu2020xtreme}.
For hate speech detection in particular, cross-lingual performance in zero-shot settings is lacking \citep{stappen2020crosslingual, leite2020toxic}.
For example, zero-shot cross-lingual transfer cannot account for language-specific taboo expressions that play a key role in classification \citep{nozza2021exposing}.
Conversely, hate speech detection models trained or fine-tuned directly on the target language, i.e. in few- and many-shot settings, are consistently found to perform best \citep{aluru2020multi, pelicon2021investigating}.

So, \textbf{how do we build hate speech detection models for hundreds more languages?}
We need at least some labelled data in the target language to make models effective, but data annotation is difficult, time-consuming and expensive.
It requires resources that are often very limited for non-English languages.
Annotating \textit{hateful} content in particular also risks exposing annotators to harm in the process \citep{vidgen2019challenges,derczynski2022handling}.

In this article, we explore strategies for hate speech detection in under-resourced languages that make \textit{efficient} use of labelled data, to build \textit{effective} models while also minimising annotation cost and risk of harm to annotators.
For this purpose, we conduct a series of experiments using mono- and multilingual models fine-tuned on differently-sized random samples of labelled hate speech data in English as well as Arabic, Spanish, Hindi, Portuguese and Italian.
Our \textbf{key findings} are:
\begin{enumerate}
    \vspace{-0.1cm}\item A small amount of labelled target-language data is needed to achieve strong performance on held-out test sets.
    \vspace{-0.1cm}\item The benefits of using more such data decrease exponentially.
    \vspace{-0.1cm}\item Initial fine-tuning on readily-available English data can partially substitute target-language data and improve model generalisability.
\end{enumerate}

Based on these findings, we formulate and discuss \textbf{five recommendations} for expanding hate speech detection into under-resourced languages:
\begin{enumerate}
    \vspace{-0.1cm}\item Collect and label target-language data.
    \vspace{-0.1cm}\item Start by labelling a small set, then iterate.
    \vspace{-0.1cm}\item Use diverse data collection methods to increase marginal benefits of annotation.
    \vspace{-0.1cm}\item Use multilingual models for unlocking readily-available data in high-resource languages for initial fine-tuning.
    \vspace{-0.1cm}\item Evaluate out-of-domain performance to reveal potential weaknesses in generalisability.
\end{enumerate}
%In short, these are to 1) collect target-language data, but 2) not to annotate much data collected using only one method and, conversely, to 3) use diverse data collection methods to make annotation more worthwhile; to 4) use multilingual models for unlocking readily-available data in high-resource languages for initial fine-tuning, and to 5) evaluate out-of-domain model performance to reveal potentially critical weaknesses in generalisability.

With these recommendations, we hope to facilitate the development of new hate speech detection models for yet-unserved languages.\footnote{We make all data and code to reproduce our experiments available on \href{https://github.com/paul-rottger/efficient-low-resource-hate-detection}{GitHub}.}

\paragraph{Definition of Hate Speech}
Definitions of hate speech vary across cultural and legal settings.
Following \citet{rottger2021hatecheck}, we define hate speech as \textit{abuse that is targeted at a protected group or at its members for being a part of that group}.
Protected groups are based on characteristics such as gender identity, race or religion, which broadly reflects Western legal consensus, particularly the US 1964 Civil Rights Act, the UK's 2010 Equality Act and the EU's Charter of Fundamental Rights.
Based on these definitions, we approach hate speech detection as the binary classification of content as either hateful or non-hateful.

%%%%%%%%%%%%%%%%%%%%%%%%%%%%%%%%%%%%%%%%%%%%%%%%%%%%%%%%%%%%%%%%%%%%%%%%%%%%%%%%%%%%
%%%%%%%%%%%%%%%%%%%%%%%%%%%%%%%%%%%%%%%%%%%%%%%%%%%%%%%%%%%%%%%%%%%%%%%%%%%%%%%%%%%%
\section{Experiments}

All our experiments follow the setup described in Figure~\ref{fig: overview}.
We start by loading a pre-trained mono- or multilingual transformer model for sequence classification.
For multilingual models, there is an optional first phase of fine-tuning on English data.
This is to simulate using readily-available data from a high-resource language that is not the target language.
For all models, there is then a second phase of fine-tuning on differently-sized random samples of data in the target language.
This is to simulate using scarce data from an under-resourced language.
Finally, all models are evaluated on the held-out test set corresponding to the target-language dataset they were fine-tuned on as well as the target-language test suite from Multilingual HateCheck \citep{rottger2022mhc}.

%%%%%%%%%%%%%%%%%%%%%%%%%%%%%%%%%%%%%%%%%%%%%%%%%%%%%%%%%%%%%%%%%%%%
\subsection{Data}

For all our experiments, we use hate speech datasets from \href{https://hatespeechdata.com/}{hatespeechdata.com}, which was first introduced by \citet{vidgen2020directions} and is now the largest public repository of datasets annotated for hate, abuse and offensive language.
At the time of our review in May 2022, the site listed 53 English datasets as well as 57 datasets in 24 other languages.
From these datasets, we select those for our experiments that
a) contain explicit labels for hate, and
b) use a definition of hate for data annotation that aligns with our own (\S\ref{sec: intro}).
% heuristic: the dataset is sufficiently large to be used in fine-tuning and evaluation -- at least 20,000 entries for FinT1 and at least 3,000 entries for FinT2.
% heuristic: the proportion of hate is sufficiently high

\subsubsection{Fine-Tuning 1: English}
For the optional first phase of fine-tuning in English, we use one of three English datasets.
%so we can evaluate the impact of English dataset choice.
\textsc{Dyn21\_En} by \citet{vidgen2021learning} contains 41,255 entries, of which 53.9\% are labelled as hateful. The entries were hand-crafted by annotators to be challenging to hate speech detection models, using the Dynabench platform \citep{kiela2021dynabench}.
\textsc{Fou18\_En} by \citet{founta2018large} contains 99,996 tweets, of which 4.97\% are labelled as hateful.
\textsc{Ken20\_En} by \citet{kennedy2020contextualizing} contains 39,565 comments from Youtube, Twitter and Reddit, of which 29.31\% are labelled as hateful.

From each of these three English datasets, we sample 20,000 entries for the optional first phase of fine-tuning, plus another 500 entries for development and 2,000 for testing.
To align the proportion of hate across English datasets, we use random sampling for \textsc{Dyn21\_En}, while for \textsc{Ken20\_En} and \textsc{Fou18\_En} we retain all hateful entries and then sample from non-hateful entries, so that the proportion of hate in the two datasets increases to 50.0\% and 22.0\%, respectively.
%, and thus better align them with the class balance in \textsc{Dyn21\_En}.

\subsubsection{Fine-Tuning 2: Target Language}
For fine-tuning in the target language, we use one of five datasets in five different target languages.
\textsc{Bas19\_Es} compiled by \citet{basile2019semeval} for SemEval 2019 contains 4,950 Spanish tweets, of which 41.5\% are labelled as hateful.
\textsc{For19\_Pt} by \citet{fortuna2019hierarchically} contains 5,670 Portuguese tweets, of which 31.5\% are labelled as hateful.
\textsc{Has21\_Hi} compiled by \citet{mandl2021hasoc} for HASOC 2021 contains 4,594 Hindi tweets, of which 12.3\% are labelled as hateful.
\textsc{Ous19\_Ar} by \citet{ousidhoum2019multilingual} contains 3,353 Arabic tweets, of which 22.5\% are labelled as hateful.
\textsc{San20\_It} compiled by \citet{sanguinetti2020haspeede} for EvalIta 2020 contains 8,100 Italian tweets, of which 41.8\% are labelled as hateful.

From each of these five target-language datasets, we randomly sample differently-sized subsets for target-language fine-tuning.
Like in English, we set aside 500 entries for development and 2,000 for testing.\footnote{Due to limited dataset size, we only set aside 300 dev and 1,000 test entries for \textsc{Ous19\_Ar} (n=3,353).}
From the remaining data, we sample subsets in 12 different sizes -- 10, 20, 30, 40, 50, 100, 200, 300, 400, 500, 1,000 and 2,000 entries -- so that we are able to evaluate the effects of using more or less labelled data within and across different orders of magnitude.\footnote{There is at least one hateful entry in every sample.}
\citet{zhao2021closer} show that there can be large sampling effects when fine-tuning on small amounts of data.
To mitigate this issue, we use 10 different random seeds for each sample size, so that in total we have 120 different samples in each language, and 600 samples across the five non-English languages.

%%%%%%%%%%%%%%%%%%%%%%%%%%%%%%%%%%%%%%%%%%%%%%%%%%%%%%%%%%%%%%%%%%%%
\subsection{Models} \label{subsec: models}

\paragraph{Multilingual Models}
We fine-tune and evaluate XLM-T \citep{barbieri2022xlmt}, an XLM-R model \citep{conneau2020xlmr} pre-trained on an additional 198 million Twitter posts in over 30 languages.
XLM-R is a widely-used architecture for multilingual language modelling, which has been shown to achieve near state-of-the-art performance on multilingual hate speech detection \citep{banerjee2021exploring,mandl2021hasoc}.
We chose XLM-T because it strongly outperformed XLM-R across our target language test sets in initial experiments.

\paragraph{Monolingual Models}
For each of the five target languages, we also fine-tune and evaluate a monolingual transformer model from HuggingFace.
For Spanish, we use \href{https://huggingface.co/pysentimiento/robertuito-base-uncased}{RoBERTuito} \citep{perez2021robertuito}.
For Portuguese, we use \href{https://huggingface.co/neuralmind/bert-base-portuguese-cased}{BERTimbau} \citep{souza2020bertimbau}.
For Hindi, we use \href{https://huggingface.co/neuralspace-reverie/indic-transformers-hi-bert}{Hindi BERT}.
For Arabic, we use \href{https://huggingface.co/aubmindlab/bert-base-arabertv02}{AraBERT v2} \citep{antoun2020arabert}.
For Italian, we use \href{https://huggingface.co/Musixmatch/umberto-commoncrawl-cased-v1}{UmBERTo}.
Details on model training can be found in Appendix \ref{app: model_training}.

\paragraph{Model Notation} 
We denote all models by an additive code.
The first part is either M for a monolingual model or X for XLM-T.
For XLM-T, the second part of the code is \textsc{DEn}, \textsc{FEn} or \textsc{KEn}, for models fine-tuned on 20,000 entries from \textsc{Dyn21\_En}, \textsc{Fou18\_En} or \textsc{Ken20\_En}.
For all models, the final part of the code is \textsc{Es}, \textsc{Pt}, \textsc{Hi}, \textsc{Ar} or \textsc{It}, corresponding to the target language that the model was finetuned on.
For example, \textsc{M+It} denotes the monolingual Italian model, UmBERTo, fine-tuned on \textsc{San20\_It}, and \textsc{X+KEn+Ar} denotes an XLM-T model fine-tuned first on 20,000 English entries from \textsc{Ken20\_En} and then on \textsc{Ous19\_Ar}.

%%%%%%%%%%%%%%%%%%%%%%%%%%%%%%%%%%%%%%%%%%%%%%%%%%%%%%%%%%%%%%%%%%%%
\subsection{Evaluation Setup} \label{subsec: evaluation}

\paragraph{Held-Out Test Sets + MHC}
We test all models on the held-out test sets corresponding to their target-language fine-tuning data, to evaluate their in-domain performance (\S\ref{subsec: held-out testsets}).
For example, we test \textsc{X+KEn+It}, which was fine-tuned on \textsc{San20\_It} data, on the \textsc{San20\_It} test set.
Additionally, we test all models on the matching target-language test suite from Multilingual HateCheck (MHC).
MHC is a collection of around 3,000 test cases for different kinds of hate as well as challenging non-hate in each of ten different languages \citep{rottger2022mhc}.
We use MHC to evaluate out-of-domain generalisability (\S\ref{subsec: mhc}).
%For example, we test \textsc{X+Ar}, which was fine-tuned on \textsc{Ous19\_Ar} data, on Arabic HateCheck.

\paragraph{Evaluation Metrics}
%The held-out test sets in each target language are imbalanced, with hate as the minority class.
%MHC, too, is imbalanced, with hate in the majority.
We use macro F1 to evaluate model performance because most of our test sets as well as MHC are imbalanced.
To give context for interpreting performance, we show baseline model results in all figures: macro F1 for always predicting the hateful class ("always hate"), for never predicting the hateful class ("never hate") and for predicting both classes with equal probability ("50/50").
We also show bootstrapped 95\% confidence intervals around the average macro F1, which is calculated across the 10 random seeds for each sample size.
These confidence intervals are expected to be wider for models fine-tuned on less data because of larger sampling effects.

\begin{figure*}[h]
\centering
\includegraphics[width=\textwidth]{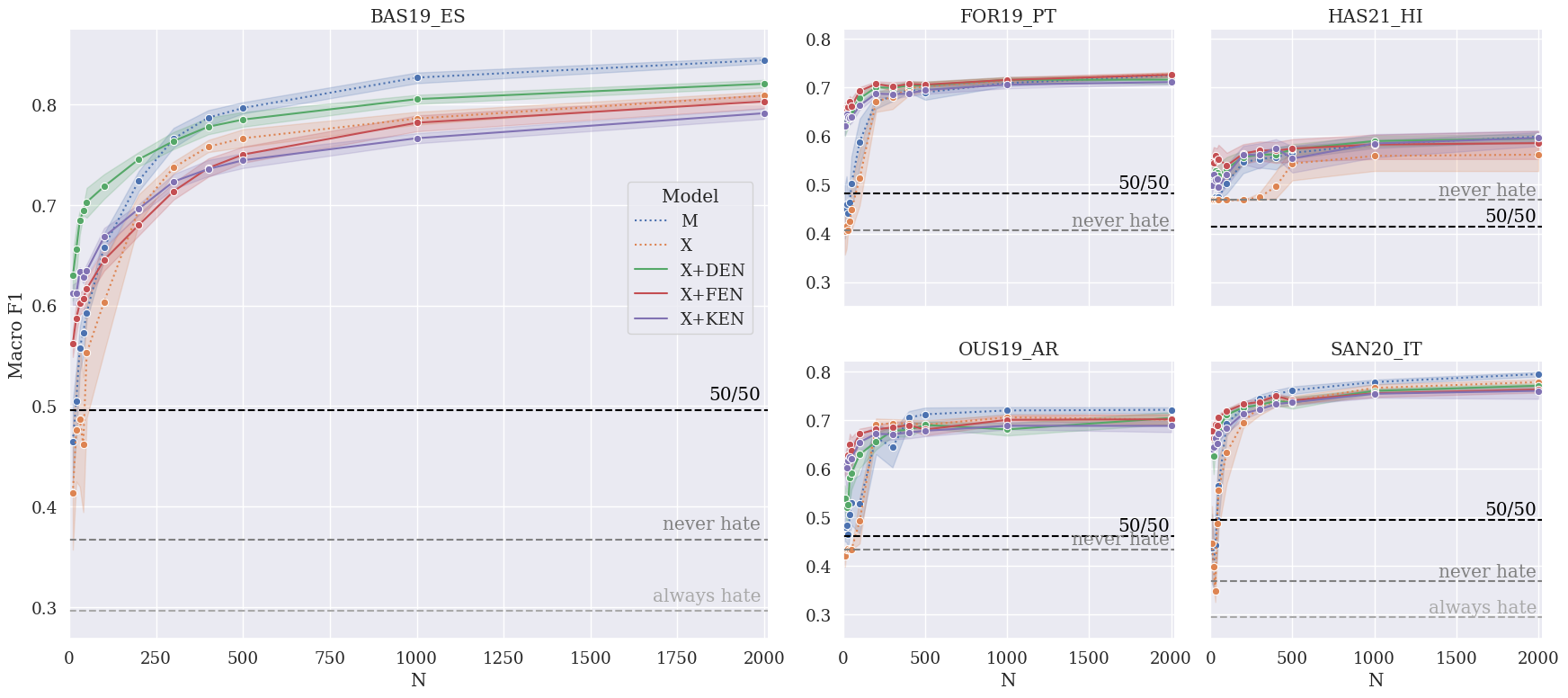}
\caption{
Macro F1 on target-language held-out test sets across models fine-tuned on up to N=2,000 target-language entries.
Model notation as described in \S\ref{subsec: models}.
Confidence intervals and model baselines as described in \S\ref{subsec: held-out testsets}.
We provide larger versions of all five graphs in Appendix \ref{app: results_testsets_full}.
}
\label{fig: macroF1_testsets_2000}
\end{figure*}

\setlength{\tabcolsep}{4.5pt}
\begin{table*}[h]
    \centering
    \resizebox{\textwidth}{!}{ 
    \begin{tabular}{l|ccc|ccc|ccc|ccc|ccc}
        \multicolumn{1}{c}{} & \multicolumn{3}{c}{\textsc{Bas19\_Es}} & \multicolumn{3}{c}{\textsc{For19\_Pt}} & \multicolumn{3}{c}{\textsc{Has21\_Hi}} & \multicolumn{3}{c}{\textsc{Ous19\_Ar}} & \multicolumn{3}{c}{\textsc{San20\_It}} \\
        \toprule
        \multicolumn{1}{r|}{\textbf{N}} & \textbf{20} & \textbf{200} & \textbf{2,000} & \textbf{20} & \textbf{200} & \textbf{2,000} & \textbf{20} & \textbf{200} & \textbf{2,000} & \textbf{20} & \textbf{200} & \textbf{2,000} & \textbf{20} & \textbf{200} & \textbf{2,000}\\
        \midrule
        \textbf{M} & 0.50 & 0.72 & \textbf{0.84} & 0.46 & 0.67 & \textbf{0.73} & 0.47 & 0.55 & \textbf{0.60} & 0.48 & 0.66 & \textbf{0.72} & 0.40 & \textbf{0.73} & \textbf{0.79} \\
        \textbf{X} & 0.48 & 0.70 & 0.81 & 0.42 & 0.67 & \textbf{0.73} & 0.47 & 0.47 & 0.56 & 0.43 & \textbf{0.69} & 0.70 & 0.40 & 0.70 & 0.78 \\
        \textbf{\textsc{X+DEn}} & \textbf{0.66} & \textbf{0.75} & 0.82 & 0.63 & 0.70 & 0.72 & 0.52 & \textbf{0.56} & \textbf{0.60 }& 0.52 & 0.66 & 0.70 & 0.63 & \textbf{0.73} & 0.77 \\
        \textbf{\textsc{X+FEn}} & 0.59 & 0.68 & 0.80 & \textbf{0.66} & \textbf{0.71} & \textbf{0.73} & \textbf{0.55} & \textbf{0.56} & 0.59 & \textbf{0.61} & 0.68 & 0.70 & \textbf{0.66} & \textbf{0.73} & 0.76 \\
        \textbf{\textsc{X+KEn}} & 0.61 & 0.70 & 0.79 & 0.65 & 0.69 & 0.71 & 0.52 & \textbf{0.56} & \textbf{0.60} & 0.60 & 0.67 & 0.69 & 0.64 & 0.71 & 0.76 \\
        \bottomrule
    \end{tabular}
    }
\caption{Macro F1 on respective held-out test sets for models fine-tuned on N target-language entries, averaged across 10 random seeds for each N. Best performance for a given N in \textbf{bold}. Results across all N in Appendix \ref{app: results_testsets_full}.}
\label{tab: macroF1_testsets_2000}
\end{table*}

%%%%%%%%%%%%%%%%%%%%%%%%%%%%%%%%%%%%%%%%%%%%%%%%%%%%%%%%%%%%%%%%%%%%
\subsection{Testing on Held-Out Test Sets} \label{subsec: held-out testsets}

When evaluating our mono- and multilingual models on their corresponding target-language test sets, we find a set of consistent patterns in model performance.
We visualise overall performance in Figure~\ref{fig: macroF1_testsets_2000} and highlight key data points in Table~\ref{tab: macroF1_testsets_2000}.

First, there is an enormous benefit from even very small amounts of target-language fine-tuning data N.
Model performance increases sharply across models and held-out test sets up to around N=200.
For example, the performance of \textsc{X+DEn+Es} increases from around 0.63 at N=10 to 0.70 at N=50, to 0.75 at N=200.

On the other hand, larger amounts of target-language fine-tuning data correspond to much less of an improvement in model performance.
Across models and held-out test sets, there is a steep decrease in the marginal benefits of increasing N.
\textsc{M+It}, for example, improves by 0.33 macro F1 from N=20 to N=200, and by just 0.06 from N=200 to N=2,000.
\textsc{X+Pt} improves by 0.25 macro F1 from N=20 to N=200, and by just 0.06 from N=200 to N=2,000.
We analyse these decreasing marginal benefits using linear regression in \S \ref{subsec: regression}.

Further, there is a clear benefit to a first phase of fine-tuning on English data, when there is limited target-language data.
Absolute performance differs across test sets, but multilingual models with initial fine-tuning on English data (i.e. \textsc{X+DEn}, \textsc{X+FEn} and \textsc{X+KEn}) perform substantially better than those without (i.e. M and X), up to around N=200.
At N=20, there is up to 0.26 macro F1 difference between the former and the latter.
Conversely, models without initial fine-tuning on English data need substantially more target-language data to achieve the same performance.
For example, \textsc{X+DEn+Pt} at N=100 performs as well as \textsc{M+Pt} at N=300 on \textsc{For19\_Pt}.

\begin{figure*}[!b]
\centering
\includegraphics[width=\textwidth]{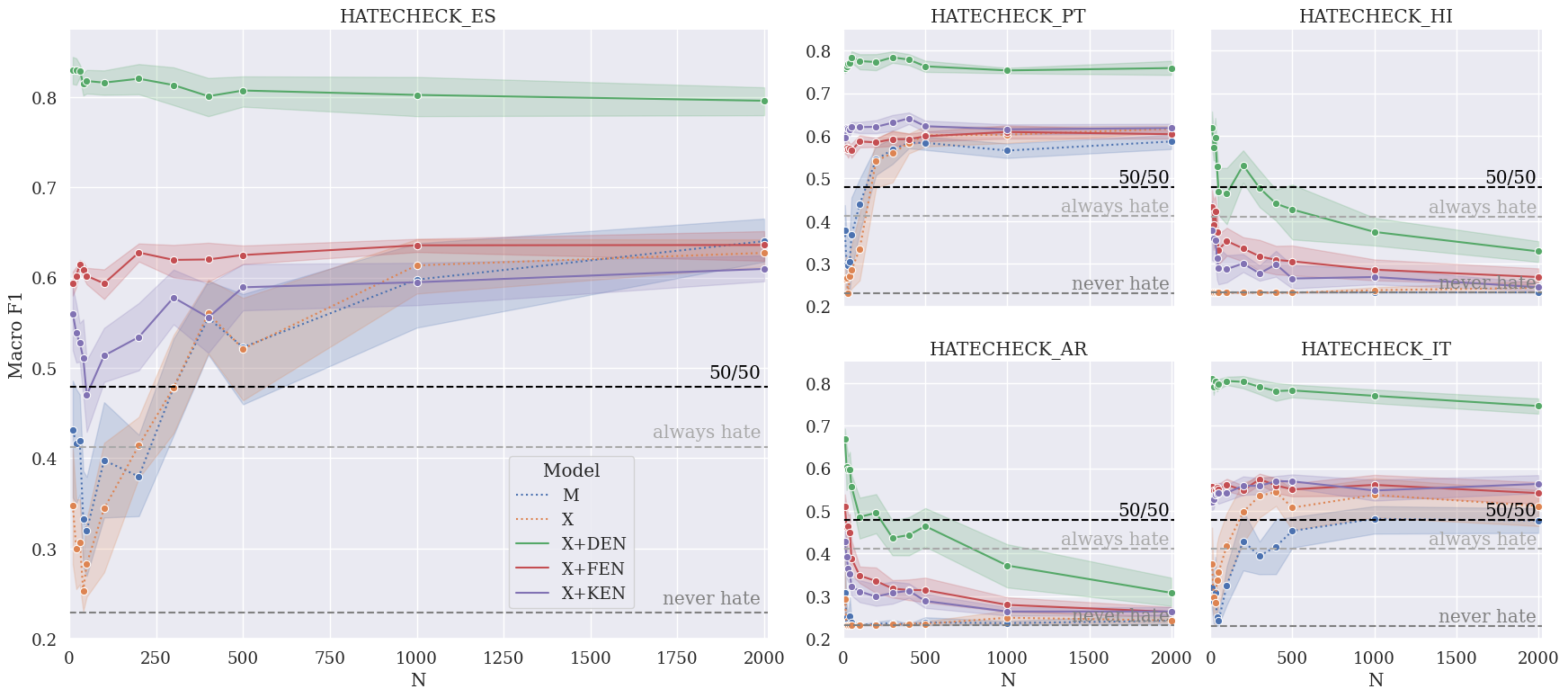}
\caption{
Macro F1 on target-language tests from MHC across models fine-tuned on up to N = 2,000 target-language entries.
Model notation as described in \S\ref{subsec: models}.
Confidence intervals and model baselines as described in \S\ref{subsec: held-out testsets}.
We provide larger versions of all five graphs in Appendix \ref{app: results_mhc_full}.
}
\label{fig: macroF1_hatecheck_2000}
\end{figure*}

\setlength{\tabcolsep}{4.5pt}
\begin{table*}[!b]
    \centering
    \resizebox{\textwidth}{!}{ 
    \begin{tabular}{l|ccc|ccc|ccc|ccc|ccc}
        \multicolumn{1}{c}{} & \multicolumn{3}{c}{\textsc{HateCheck\_Es}} & \multicolumn{3}{c}{\textsc{HateCheck\_Pt}} & \multicolumn{3}{c}{\textsc{HateCheck\_Hi}} & \multicolumn{3}{c}{\textsc{HateCheck\_Ar}} & \multicolumn{3}{c}{\textsc{HateCheck\_It}} \\
        \toprule
        \multicolumn{1}{r|}{\textbf{N}} & \textbf{20} & \textbf{200} & \textbf{2,000} & \textbf{20} & \textbf{200} & \textbf{2,000} & \textbf{20} & \textbf{200} & \textbf{2,000} & \textbf{20} & \textbf{200} & \textbf{2,000} & \textbf{20} & \textbf{200} & \textbf{2,000}\\
        \midrule
        \textbf{M} & 0.42 & 0.38 & 0.64 & 0.30 & 0.54 & 0.59 & 0.23 & 0.23 & 0.23 & 0.25 & 0.23 & 0.24 & 0.29 & 0.43 & 0.48 \\
        \textbf{X} & 0.30 & 0.41 & 0.63 & 0.27 & 0.54 & 0.62 & 0.23 & 0.23 & 0.24 & 0.23 & 0.23 & 0.24 & 0.30 & 0.50 & 0.51 \\
        \textbf{\textsc{X+DEn}} & \textbf{0.83} & \textbf{0.82} & \textbf{0.80} & \textbf{0.76} & \textbf{0.77} & \textbf{0.76} & \textbf{0.57} & \textbf{0.53} & \textbf{0.33} & \textbf{0.60} & \textbf{0.50} & \textbf{0.31} & \textbf{0.79} & \textbf{0.80} & \textbf{0.75} \\
        \textbf{\textsc{X+FEn}} & 0.60 & 0.63 & 0.64 & 0.57 & 0.58 & 0.60 & 0.39 & 0.34 & 0.27 & 0.46 & 0.34 & 0.26 & 0.55 & 0.55 & 0.54 \\
        \textbf{\textsc{X+KEn}} & 0.54 & 0.53 & 0.61 & 0.62 & 0.62 & 0.62 & 0.36 & 0.30 & 0.24 & 0.39 & 0.30 & 0.26 & 0.53 & 0.56 & 0.56 \\
        \bottomrule
    \end{tabular}
    }
\caption{Macro F1 on target-language HateCheck for models fine-tuned on N target-language entries, averaged across 10 random seeds for each N. Best performance for a given N in \textbf{bold}. Results across all N in Appendix \ref{app: results_mhc_full}.}
\label{tab: macroF1_hatecheck_2000}
\end{table*}

Relatedly, there are clear differences in model performance based on which English dataset was used in the first fine-tuning phase, when there is limited target-language data.
Among the three English datasets we evaluated, \textsc{X+DEn} performs best for up to around N=200 on \textsc{Bas19\_Es}, whereas \textsc{X+FEn} wins out for the other four test sets.

After the strong initial divergence for low amounts of target-language fine-tuning data, performance tends to align across models, regardless of whether they were first fine-tuned on English data or only fine-tuned on the target language.
From around N=200 onwards, for a given test set, all models we evaluate have broadly comparable performance.
For example, all models score 0.71 to 0.72 macro F1 on \textsc{For19\_Pt} for N=1,000.

Finally, we find that if there is some divergence in model performance on the held-out test sets for larger amounts of target-language fine-tuning data, then monolingual models tend to outperform multilingual models.
For example, despite the disadvantage of monolingual models for N$\leq$200, we find that \textsc{M+Es} achieves 0.84 macro F1 on \textsc{Bas19\_Es} for N=2,000, compared 0.82 macro F1 for \textsc{X+DEn+Es}.
This difference is visible across test sets and significant at 95\% confidence for the Spanish, Arabic and Italian test sets.

%Generally, the \textbf{initial performance improvements from fine-tuning on more target-language data are more pronounced for those models that were not first fine-tuned on English data}.
%Models that are fine-tuned only on target-language data (i.e. M and X) benefit the most from an increase in N up to around N=200.
%For example, the performance of \textsc{M+It} on \textsc{San20\_It} increases by 0.33 macro F1 from 0.40 at N=20 to 0.73 at N=200, whereas that of \textsc{X+KEn+It} increases by only 0.07 macro F1 from 0.64 to 0.71.

%%%%%%%%%%%%%%%%%%%%%%%%%%%%%%%%%%%%%%%%%%%%%%%%%%%%%%%%%%%%%%%%%%%%
\subsection{Testing on MHC} \label{subsec: mhc}

Performance on the target-language test suites from MHC, which none of the models saw during fine-tuning, differs strongly and systematically from results on the held-out test sets (Figure~\ref{fig: macroF1_hatecheck_2000}).
Key data points are highlighted in Table~\ref{tab: macroF1_hatecheck_2000}.

First, as for the held-out test sets, there is an initial benefit to a first phase of fine-tuning on English data, up to around N=500 for MHC.
However, there are clear performance differences depending on which English dataset was used.
Models fine-tuned on \textsc{Dyn21\_En} perform best by far across MHC for any N.
On Italian HateCheck for example, \textsc{X+DEn} achieves 0.79 macro F1 at N=20 compared to 0.55 for \textsc{X+FEn+It} and 0.29 for the monolingual model \textsc{M+It}.

Second, for Spanish, Portuguese and Italian, there is little to no benefit to fine-tuning on more target-language data, as measured by performance on MHC.
Only those models that were not first fine-tuned on English data (i.e. M and X) are improved by fine-tuning on more target-language data, up to around N=500.
None of the models benefit from even larger N.
On Portuguese HateCheck for example, \textsc{X+Ken+Pt} scores 0.62 macro F1 at N=500, N=1,000 and N=2,000.

Third, for Hindi and Arabic, training on more target-language data actively harms performance on MHC.
For larger N, all models degrade towards only predicting the non-hateful class.
On Arabic HateCheck, for example, \textsc{X+FEn+Ar} achieves 0.26 macro F1 at N=2,000, which is only slightly better than the ``never hate'' baseline at 0.23 macro F1.
This is despite the exact same model scoring 0.70 macro F1 on \textsc{Ous19\_Ar} at N=2,000.
We further discuss this finding in \S\ref{sec: discussion}.

Lastly model performance on MHC is much more volatile across random seeds compared to performance on the held-out test sets.
This can be seen from the wider confidence intervals in Figure~\ref{fig: macroF1_hatecheck_2000} compared to Figure~\ref{fig: macroF1_testsets_2000}.

%%%%%%%%%%%%%%%%%%%%%%%%%%%%%%%%%%%%%%%%%%%%%%%%%%%%%%%%%%%%%%%%%%%%
\subsection{Regression Analysis} \label{subsec: regression}

When evaluating models on their corresponding held-out test sets, we saw decreasing benefits to adding more target-language fine-tuning data (\S\ref{subsec: held-out testsets}).
To quantify this observation and evidence its significance, we fit ordinary least squares (OLS) linear regressions for each model on each test set.
OLS is commonly used in economics and the social sciences to analyse empirical relationships between an outcome variable and one or more regressor variables.
In our case, the outcome is model performance as measured by macro F1 and the regressor is the amount of target-language fine-tuning data N.
Since the relationship between the two variables is clearly non-linear and plausibly exponential, we regress macro F1 on the natural logarithm of N instead of just N, so that our regression equation for a given random seed $i$ takes the form of

$$ F_i = \beta_0 + \beta_1 ln(N_i) + \varepsilon_i $$

where $F_i$ is the observed macro F1 and $\varepsilon_i$ the error term.
$\beta_0$, the intercept, can be interpreted as the expected zero-shot macro F1, i.e. at N=0.
$\beta_1$, the slope, can be interpreted as the expected increase in macro F1 from increasing the amount of target-language fine-tuning data N by 1\%, due to our log-linear regression setup.
In addition to reporting regression coefficients, we also report adjusted $R^2$ as a goodness-of-fit measure, which is calculated as the percentage of variance in macro F1 explained by our regression.
Lastly, we report the expected percentage-point increase in macro F1 from doubling N, denoted as $2N$, which is calculated as $\beta_1 ln(2)$.
Since we saw that variance in macro F1 changes with N, we use heteroskedasticity-robust HC3 standard errors \citep{hayes2007using}.

\begin{table}[h]
    \centering
    \resizebox{0.48\textwidth}{!}{ 
    \begin{tabular}{l|cc|c|c}
        \multicolumn{5}{l}{Model: M} \\
        \toprule
        \textbf{Test Set} &  \bm{$\beta_0$} &  \bm{$\beta_1$} & \bm{$R^2$} & \bm{$2N$}\\
        \midrule
        \textbf{\textsc{Bas19\_Es}} & 0.2879 & 0.0792 & 91.97 & 5.49 \\
        \textbf{\textsc{For19\_Pt}} & 0.2628 & 0.0670 & 71.23 & 4.65 \\
        \textbf{\textsc{Has21\_Hi}} & 0.2998 & 0.0607 & 69.11 & 4.21 \\
        \textbf{\textsc{Ous19\_Ar}} & 0.2092 & 0.0866 & 70.08 & 6.00 \\
        \textbf{\textsc{San20\_It}} & 0.3784 & 0.0294 & 77.68 & 2.04 \\
        \bottomrule
        \multicolumn{5}{l}{} \\
        \multicolumn{5}{l}{Model: \textsc{X+DEn}} \\
        \toprule
        \textbf{Test Set} &  \bm{$\beta_0$} &  \bm{$\beta_1$} & \bm{$R^2$} & \bm{$2N$}\\
        \midrule
        \textbf{\textsc{Bas19\_Es}} & 0.5543 & 0.0363 & 87.75 & 2.52 \\
        \textbf{\textsc{For19\_Pt}} & 0.5756 & 0.0205 & 63.66 & 1.42 \\
        \textbf{\textsc{Has21\_Hi}} & 0.4691 & 0.0163 & 43.91 & 1.13 \\
        \textbf{\textsc{Ous19\_Ar}} & 0.4320 & 0.0391 & 64.99 & 2.71 \\
        \textbf{\textsc{San20\_It}} & 0.5724 & 0.0272 & 69.42 & 1.89 \\
        \bottomrule
        
    \end{tabular}
    }
\caption{OLS results for \textsc{M} and \textsc{X+DEn}. All regression coefficients $\beta_0$ and  $\beta_1$ are significant at $p<0.001$.
We show results for all models in Appendix \ref{app: ols_full}.}
\label{tab: ols}
\end{table}

We find that the natural logarithm of the amount of target-language fine-tuning data N provides a strikingly good explanation for model performance as measured by macro F1 (Table~\ref{tab: ols}).
$R^2$ for our regressions, where N is the sole explanatory variable, is generally very high, up to 91.97\%, which indicates near-perfect goodness-of-fit.
This strongly suggests that there is an exponential decrease in the benefits of using more target-language fine-tuning data.
Conversely, there is a constant benefit to doubling the amount of target-language fine-tuning data: 
We expect the same increase in macro F1 whether we increase N ten-fold from 10 to 20 or from 1,000 to 2,000.

Further, we can confirm our earlier findings that a) initial fine-tuning on English data improves performance for low N, while b) all models perform similarly well for larger N.
In our regression, this is visible in monolingual M models having much lower estimated zero-shot performance $\beta_0$ paired with much larger $\beta_1$ and $2N$ effects compared to the \textsc{X+DEn} models.
For example, doubling the amount of Spanish fine-tuning data for M ($\beta_0$=0.29) corresponds to an expected increase in macro F1 of 5.49 points on \textsc{Bas19\_Es}, whereas for \textsc{X+DEn} ($\beta_0$=0.55) we only expect an increase of 2.52 points.

%%%%%%%%%%%%%%%%%%%%%%%%%%%%%%%%%%%%%%%%%%%%%%%%%%%%%%%%%%%%%%%%%%%%%%%%%%%%%%%%%%%%
%%%%%%%%%%%%%%%%%%%%%%%%%%%%%%%%%%%%%%%%%%%%%%%%%%%%%%%%%%%%%%%%%%%%%%%%%%%%%%%%%%%%
\section{Discussion and Recommendations} \label{sec: discussion}

Based on the sum of our experimental results, we formulate five recommendations for the data-efficient development of effective hate speech detection models for under-resourced languages.

\begin{RecBox}{Recommendation 1}
Collect and label target-language data.
\end{RecBox}
We found enormous benefits from fine-tuning models on even small amounts of target-language data.
Zero-shot cross-lingual transfer performs worse than naive baselines, but  few shots go a long way.

\begin{RecBox}{Recommendation 2}
Do not annotate all the data you collect right away.
Instead, start small and iterate.
\end{RecBox}
We found that the benefits of using more target-language fine-tuning data decrease exponentially (Table~\ref{tab: ols}), which suggests that the data in each dataset is not diverse enough to warrant full annotation.
The datasets we used in our experiments contained several thousand entries, but most of the useful signal in them can be learned from just a few dozen or hundred randomly-sampled entries.\footnote{Future work could explore more deliberate selection strategies, like active learning, which would likely achieve even better performance with fewer target-language entries \citep[see for example][]{markov2022holistic}.}
Thus, most of the annotation cost and the potential harm to annotators in compiling these datasets created very little performance benefit.
Assuming constant cost per additional annotation, improving F1 by just one percentage point grows exponentially more expensive.
Our results suggest that dataset creators should at first annotate smaller subsets of the data they collected and iteratively evaluate what marginal benefits additional annotations are likely to bring compared to their costs.

\begin{RecBox}{Recommendation 3}
Use diverse data collection strategies to minimise redundant data annotation.
\end{RecBox}
We do not recommend against annotating target-language data.
Instead, we recommend against annotating too much data that was collected using the same method.
The \textsc{Has21\_Hi} dataset, for example, was collected using trending hashtags related to Covid and Indian politics \citep{mandl2021hasoc}, and \citet{ousidhoum2019multilingual} used topical keywords related to Islamic inter-sect hate for \textsc{Ous19\_Ar}.
As a consequence, these datasets necessarily have a narrow topical focus as well as limited linguistic diversity.
This kind of \textit{selection bias} has been confirmed for several other hate speech datasets \citep{wiegand2019detection,ousidhoum2020comparative}.
MHC, on the other hand, contains diverse statements against seven different target groups in each language \citep{rottger2022mhc}.
Our results show that very quickly additional entries from a given dataset carry little new and useful information (Figure~\ref{fig: macroF1_testsets_2000}), and that fine-tuning on particularly narrow datasets like \textsc{Has21\_Hi} and \textsc{Ous19\_Ar} can even harm model generalisability as measured by MHC (Figure~\ref{fig: macroF1_hatecheck_2000}).
Conversely, more diverse strategies for data collection, such as sampling from multiple platforms, sampling from a diverse set of users, or using a wider variety of topical keywords, are likely to result in data that is more worth annotating.

\begin{RecBox}{Recommendation 4}
Use multilingual models to unlock readily-available data in high-resource languages for initial fine-tuning.
\end{RecBox}
We found that an initial phase of fine-tuning on English data increases model performance on held-out test sets when there is little target-language fine-tuning data (Figure~\ref{fig: macroF1_testsets_2000}).
Thus, English data can partly substitute target-language data in few-shot settings.
It also substantially improves out-of-domain generalisability to the MHC test suites (Figure~\ref{fig: macroF1_hatecheck_2000}), with benefits varying by which English dataset is used.
Only multilingual models can access these benefits, because they can be fine-tuned on languages other than just the target language.
In our experiments, we used only three of the many English hate speech datasets that have been published to date.
Of them, we use only one at a time, and only a subset of the full data.
Future work could explore the benefits from initial fine-tuning on larger and more diverse datasets, or a combination of datasets in English and other languages that may be more available than the target language.

\begin{RecBox}{Recommendation 5}
Do not rely on just one held-out test set to evaluate model performance in the target language.
\end{RecBox}
Monolingual models appear strong when evaluated on held-out test sets, and sometimes even outperform multilingual models when there is relatively much target-language fine-tuning data (Figure~\ref{fig: macroF1_testsets_2000}).
However, this hides clear weaknesses in their generalisability as measured by MHC (Figure~\ref{fig: macroF1_hatecheck_2000}).
These findings suggest that monolingual models are more at risk of overfitting to dataset-specific features than multilingual models fine-tuned first on data from another language.
In higher-resource settings, monolingual models are still likely to outperform multilingual models \citep{martin2020camembert,nozza2020what,armengol2021multilingual}.

%%%%%%%%%%%%%%%%%%%%%%%%%%%%%%%%%%%%%%%%%%%%%%%%%%%%%%%%%%%%%%%%%%%%%%%%%%%%%%%%%%%%
%%%%%%%%%%%%%%%%%%%%%%%%%%%%%%%%%%%%%%%%%%%%%%%%%%%%%%%%%%%%%%%%%%%%%%%%%%%%%%%%%%%%
\section{Related Work} \label{sec: related work}

\paragraph{Cross-Lingual Transfer Learning}
Only a very small number of the over 7,000 languages in the world are well-represented in natural language processing resources \citep{joshi2020state}.
This has motivated much research into cross-lingual transfer, where large multilingual language models are fine-tuned on one (usually high-resource) source language and then applied to another (usually under-resourced) target language.
Cross-lingual transfer has been systematically evaluated for a wide range of NLP tasks and for varying amounts of \textit{shots}, i.e. target-language training data.
Evidence on the efficacy of \textit{zero-shot} transfer is mixed, across tasks and for transfer between different language pairs \citep[][inter alia]{artetxe2019massively,pires2019multilingual,wu2019beto,lin2019choosing,hu2020xtreme, liang2020xglue, ruder2021xtremer}.
%although their effectiveness is increased by fine-tuning on intermediate tasks \citep{phang2020english}
In comparison, \textit{few-shot} approaches are generally found to be more effective \citep{lauscher2020zero,hedderich2020transfer,zhao2021closer,hung2022multi2woz}.
In \textit{many-shot} settings, multilingual models are often outperformed by their language-specific monolingual counterparts \citep{martin2020camembert,nozza2020what,armengol2021multilingual}.
In this paper, we confirmed and expanded on key results from this body of prior work, such as the effectiveness of few-shot learning, specifically for the task of hate speech detection.
We also introduced OLS linear regression as an effective method for quantifying data efficiency, i.e. dynamic cost-benefit trade-offs in data annotation.

\paragraph{Cross-Lingual Hate Speech Detection}

Research on cross-lingual hate speech detection has generally had a more narrow scope than ours. 
\citet{nozza2020what} demonstrate weaknesses of zero-shot transfer between English, Italian and Spanish.
\citet{leite2020toxic} find similar results for Brazilian Portuguese, \citet{bigoulaeva2021cross} for German, and \citet{stappen2020crosslingual} for Spanish, while also providing initial evidence for decreasing returns to fine-tuning on more hate speech data \citep[see also][]{aluru2020multi}.
\citet{pelicon2021investigating} and \citet{ahn2020nlpdove}, like \citet{stappen2020crosslingual}, find that intermediate training on other languages is effective when little target-language data is available.
By comparison, we covered a wider range of languages, with two datasets per language and a consistent definition of hate across datasets.
We provided a more systematic evaluation of marginal benefits to additional target-language fine-tuning data based on OLS linear regression, as well as a first assessment of out-of-domain performance using the recently-released Multilingual HateCheck test suites.
We are also the first to formulate concrete recommendations for expanding hate speech detection into more under-resourced languages.

%%%%%%%%%%%%%%%%%%%%%%%%%%%%%%%%%%%%%%%%%%%%%%%%%%%%%%%%%%%%%%%%%%%%%%%%%%%%%%%%%%%%
%%%%%%%%%%%%%%%%%%%%%%%%%%%%%%%%%%%%%%%%%%%%%%%%%%%%%%%%%%%%%%%%%%%%%%%%%%%%%%%%%%%%
\section{Conclusion} \label{sec: conclusion}
In this paper, we explored strategies for hate speech detection in under-resourced language settings that are effective while also minimising annotation cost and risk of harm to annotators.

We conducted a series of experiments using mono- and multilingual language models fine-tuned on differently-sized random samples of labelled hate speech data in English as well as Arabic, Spanish, Hindi, Portuguese and Italian.
We evaluated all models on held-out test sets as well as out-of-domain target-language test suites from Multilingual HateCheck, and used OLS linear regression to quantify the marginal benefits of target-language fine-tuning data.
In our experiments, we found that 
1) a small amount of target-language fine-tuning data is needed to achieve strong performance on held-out test sets,
2) the benefits of using more such data decrease exponentially, and
3) initial fine-tuning on readily-available English data can partially substitute target-language data and improve model generalisability.
Based on our findings, we formulated five actionable recommendations for expanding hate speech detection into under-resourced languages.

Most hate speech research and resources so far have focused on English-language content.
However, hate speech is a global phenomenon, and hate speech detection models are needed in hundreds more languages.
With our findings and recommendations, we hope that we can facilitate the development of models for yet-unserved languages and thus help to close language gaps that right now leave billions of non-English speakers world less protected against online hate.

\vspace{0.6cm}
\hrule
\vspace{0.1cm}
\hrule
\vspace{0.3cm}

%%%%%%%%%%%%%%%%%%%%%%%%%%%%%%%%%%%%%%%%%%%%%%%%%%%%%%%%%%%%%%%%%%%%
\section*{Acknowledgments}
%PR
This article was completed during Paul Röttger's 2022 research visit to Dirk Hovy's research group at Bocconi University in Milan.
Paul Röttger was funded by the German Academic Scholarship Foundation.
%DN
Debora Nozza received funding from Fondazione Cariplo (grant No. 2020-4288, MONICA).
%FB
Federico Bianchi was a member of the Bocconi Institute for Data Science and Analysis (BIDSA) when parts of this work were completed.
%DH
Dirk Hovy received funding from the European Research Council (ERC) under the European Union’s Horizon 2020 research and innovation program (grant agreement No.\ 949944).
Debora Nozza and Dirk Hovy are members of the Data and Marketing Insights (DMI) Unit of the Bocconi Institute for Data Science and Analysis (BIDSA).
% General
We thank the MilaNLP Lab and the Pierrehumbert Language Modelling Group for their helpful comments, and all reviewers for their constructive feedback.

\vspace{0.6cm}
\hrule
\vspace{0.1cm}
\hrule
\vspace{0.3cm}

%%%%%%%%%%%%%%%%%%%%%%%%%%%%%%%%%%%%%%%%%%%%%%%%%%%%%%%%%%%%%%%%%%%%
\section*{Limitations}

Our analysis is necessarily constrained by the availability of suitable hate speech datasets, particularly for non-English languages.
The datasets we used are idiosyncratic, not just because their language differs but also because they were collected using different methods at different times.
This is why we focus on analysing general patterns across datasets in different languages, rather than attempting like-for-like comparisons between individual languages.
As new datasets are created, future work could expand our analysis to cover more languages as well as more datasets within each language.

Further, our experiments and the recommendations we formulate based on their results presuppose the availability of at least some models tailored to the target language.
Multilingual models like XLM-R and XLM-T are respectively pre-trained on 100 and on 30 languages \citep{conneau2020xlmr,barbieri2022xlmt}, and there are active efforts to expand language model coverage to hundreds of more languages \citep{wang2022expanding}.
However, even the most multilingual models available today cover only a fraction of the world's languages.
Similarly, there is a growing number of monolingual transformer models for different languages \citep{nozza2020what} but most languages are missing such a model.

Also, we focus on varying amounts of fine-tuning data while assuming the existence of held-out development and test sets to evaluate our experiments.
However large or small these sets are, labelling them also creates costs and risks of harm to annotators.
This needs to be factored into cost assessments for expanding hate speech detection into an under-resourced language.

For reasons of scope and prioritisation, we did not explore the full range of possibilities for improving cross-lingual model performance.
For example, we chose English data for the initial fine-tuning because it is most readily available, but English may not be the optimal choice \citep{lin2019choosing}.
Further, prior work on hate speech detection has found some benefits to data augmentation based on machine translation \citep{pamungkas2021joint, wang2021practical} or semi-supervised learning \citep{zia2022improving}, which could be explored in future research.

Lastly, our definition of hate speech (\S\ref{sec: intro}) is based on a Western legal consensus, in which the presence of hate speech hinges on the target being a protected group.
This definition is consistent throughout the datasets we use for our experiments.
However, which groups are considered protected may differ across legal and cultural settings.
The United Arab Emirates, for example, do not consider sexual orientation a protected characteristic \citep{uae2022hate}.
To avoid Eurocentrism, potential differences across legal and cultural settings need to be accounted for when expanding hate speech detection into new languages and when creating new hate speech datasets.

%%%%%%%%%%%%%%%%%%%%%%%%%%%%%%%%%%%%%%%%%%%%%%%%%%%%%%%%%%%%%%%%%%%%
\section*{Ethical Considerations}

\paragraph{Language Representation}
Besides English, our experiments cover a limited set of five languages: Arabic, Spanish, Hindi, Portuguese and Italian.
This is due to the very scarcity of suitable non-English datasets for hate speech detection that our work seeks to address.
While certainly under-resourced and under-represented in the context of hate speech detection, the five non-English languages we consider are widely spoken, and they have received at least some attention in hate speech research.
There is a clear need for creating hate speech datasets in even more under-resourced languages, to facilitate the creation of hate speech detection models in these languages and also to enable even more comprehensive analyses of the kind we conducted here.

\paragraph{Biases in Multilingual Models}
\citet{zhao2020gender} show that multilingual models exhibit similar biases as their monolingual counterparts.
Therefore, there is a risk that the models we trained, especially those trained on fewer target-language entries, reflect the biases present in their training and fine-tuning data.
Our analyses focused on general model efficacy rather than bias evaluation.
Future work could for example make use of Multilingual HateCheck to evaluate model performance across different targeted groups.

\paragraph{Environmental Impact}
We fine-tuned 3,000 models for our experiments: five model types for each of five target languages with 12 differently-sized samples for 10 random seeds in each language.
However, each model fine-tuning step was very computationally inexpensive, especially because we used very small amounts of fine-tuning data.
We were able to run all our experimentation on the University of Oxford's Advanced Research Computing cloud CPUs.
Fine-tuning 120 models for a given language took only around four hours for monolingual models and around seven hours for XLM-T. 
Relative to the concerns raised around the environmental costs of pre-training large language models \citep{strubell2019energy,henderson2020towards,bender2021stochastic}, or even larger-scale fine-tuning with hyperparameter tuning, we therefore consider the environmental costs of our work to be relatively minor.
Further, our findings enable the training more effective models with less data and may thus reduce the environmental impact of future work.

%%%%%%%%%%%%%%%%%%%%%%%%%%%%%%%%%%%%%%%%%%%%%%%%%%%%%%%%%%%%%%%%%%%%%%%%%%%%%%%%%%%%
%%%%%%%%%%%%%%%%%%%%%%%%%%%%%%%%%%%%%%%%%%%%%%%%%%%%%%%%%%%%%%%%%%%%%%%%%%%%%%%%%%%%

\bibliography{custom}
\bibliographystyle{acl_natbib}

%\clearpage
\appendix
%%%%%%%%%%%%%%%%%%%%%%%%%%%%%%%%%%%%%%%%%%%%%%%%%%%%%%%%%%%%%%%%%%%%%%%%%%%%%%%%%%%%
%%%%%%%%%%%%%%%%%%%%%%%%%%%%%%%%%%%%%%%%%%%%%%%%%%%%%%%%%%%%%%%%%%%%%%%%%%%%%%%%%%%%
\section{Details on Model Training} \label{app: model_training}

\paragraph{Parameters}
We implemented XLM-T as well as the five language-specific monolingual models (\S\ref{subsec: models}) using the HuggingFace \texttt{transformers} library \citep{wolf2020transformers}.
Training batch size was 16.
Maximum sequence length was 128 tokens.
Otherwise, we used default parameters.
The optional first phase of fine-tuning on English data was for three epochs.
Fine-tuning on the target language was for five epochs, with subsequent best epoch selection based on macro F1 on the held-out development set.

\paragraph{Computation}
We ran all experiments on our institution's 16-core cloud CPUs.
Fine-tuning 120 models for a one of the five target languages took around four hours for monolingual models and around seven hours for XLM-T.

%%%%%%%%%%%%%%%%%%%%%%%%%%%%%%%%%%%%%%%%%%%%%%%%%%%%%%%%%%%%%%%%%%%%%%%%%%%%%%%%%%%%
%%%%%%%%%%%%%%%%%%%%%%%%%%%%%%%%%%%%%%%%%%%%%%%%%%%%%%%%%%%%%%%%%%%%%%%%%%%%%%%%%%%%
\section{Macro F1 on Held-Out Test Sets} \label{app: results_testsets_full}

\begin{figure*}[h]
\centering
\includegraphics[height=0.9\textheight]{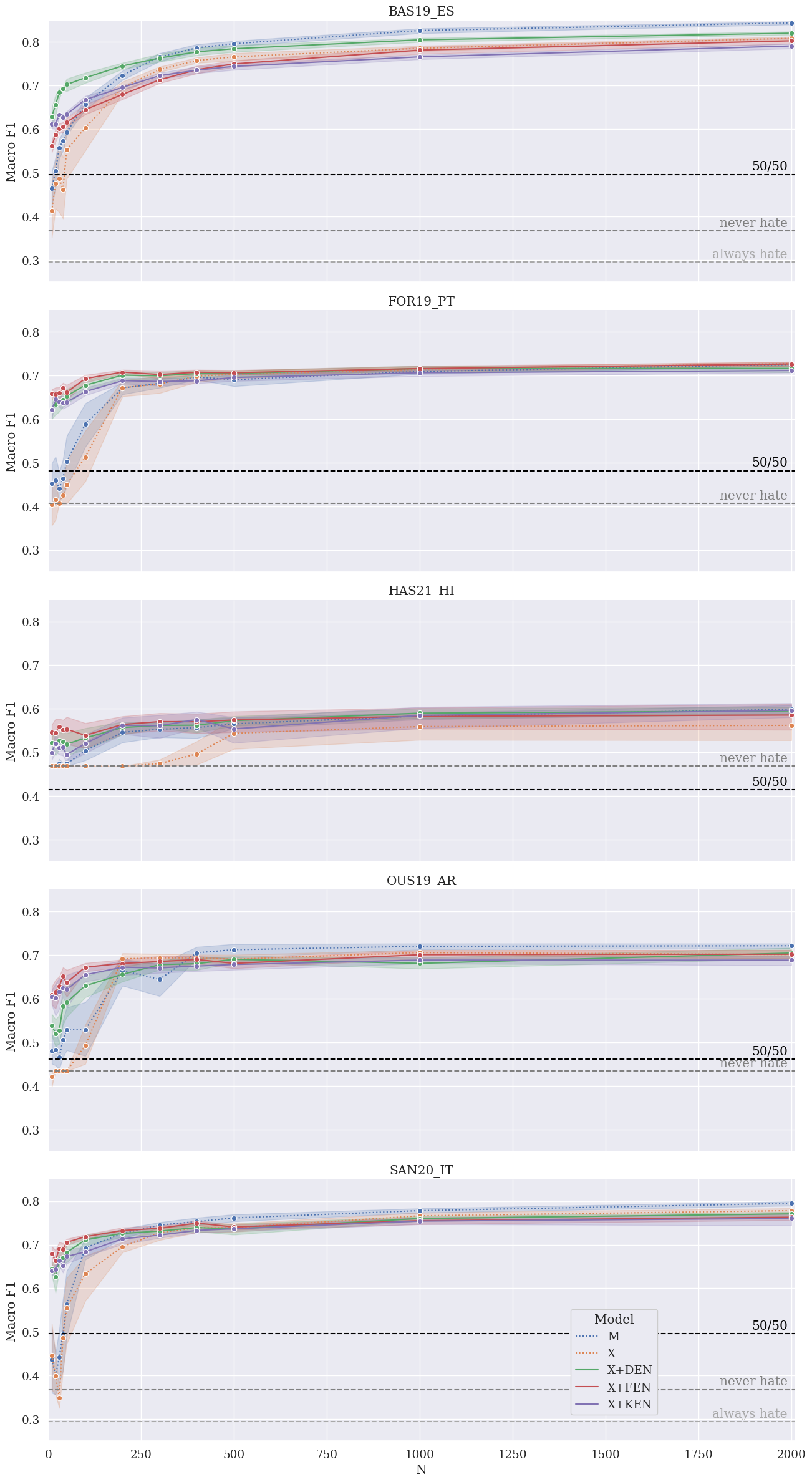}
\caption{
Macro F1 on target-language held-out test sets across models trained on up to N=2,000 target-language entries.
Model notation as described in \S\ref{subsec: models}.
Confidence intervals and model baselines as described in \S\ref{subsec: held-out testsets}.
}
\label{fig: macroF1_testsets_fullpage}
\end{figure*}

For larger versions of the graphs in Figure~\ref{fig: macroF1_testsets_2000} from the main body, see Figure~\ref{fig: macroF1_testsets_fullpage} on the next page.
See Table~\ref{tab: macroF1_testsets_fullpage} for a version of Table~\ref{tab: macroF1_testsets_2000} across all N.

%%%%%%%%%%%%%%%%%%%%%%%%%%%%%%%%%%%%%%%%%%%%%%%%%%%%%%%%%%%%%%%%%%%%%%%%%%%%%%%%%%%%
%%%%%%%%%%%%%%%%%%%%%%%%%%%%%%%%%%%%%%%%%%%%%%%%%%%%%%%%%%%%%%%%%%%%%%%%%%%%%%%%%%%%
\section{Macro F1 on MHC} \label{app: results_mhc_full}

For larger versions of the graphs in Figure~\ref{fig: macroF1_hatecheck_2000} from the main body, see Figure~\ref{fig: macroF1_hatecheck_fullpage} below.
Similarly, see Table~\ref{tab: macroF1_hatecheck_fullpage} for a version of Table~\ref{tab: macroF1_hatecheck_2000} across all N.

%%%%%%%%%%%%%%%%%%%%%%%%%%%%%%%%%%%%%%%%%%%%%%%%%%%%%%%%%%%%%%%%%%%%%%%%%%%%%%%%%%%%
%%%%%%%%%%%%%%%%%%%%%%%%%%%%%%%%%%%%%%%%%%%%%%%%%%%%%%%%%%%%%%%%%%%%%%%%%%%%%%%%%%%%
\section{Regression Results for All Models} \label{app: ols_full}
See Table \ref{tab: ols_full} to the right.

\begin{table}[!t]
    \centering
    \resizebox{0.48\textwidth}{!}{ 
    \begin{tabular}{l|cc|c|c}
        \multicolumn{5}{l}{Model: M} \\
        \toprule
        \textbf{Test Set} &  \bm{$\beta_0$} &  \bm{$\beta_1$} & \bm{$R^2$} & \bm{$2N$}\\
        \midrule
        \textbf{\textsc{Bas19\_Es}} & 0.2879 & 0.0792 & 91.97 & 5.49 \\
        \textbf{\textsc{For19\_Pt}} & 0.2628 & 0.0670 & 71.23 & 4.65 \\
        \textbf{\textsc{Has21\_Hi}} & 0.2998 & 0.0607 & 69.11 & 4.21 \\
        \textbf{\textsc{Ous19\_Es}} & 0.2092 & 0.0866 & 70.08 & 6.00 \\
        \textbf{\textsc{San20\_It}} & 0.3784 & 0.0294 & 77.68 & 2.04 \\
        \bottomrule
        \multicolumn{5}{l}{} \\
        \multicolumn{5}{l}{Model: X} \\
        \toprule
        \textbf{Test Set} &  \bm{$\beta_0$} &  \bm{$\beta_1$} & \bm{$R^2$} & \bm{$2N$}\\
        \midrule
        \textbf{\textsc{Bas19\_Es}} & 0.2107 & 0.0856 & 75.39 & 5.93 \\
        \textbf{\textsc{For19\_Pt}} & 0.1727 & 0.0808 & 79.17 & 5.6 \\
        \textbf{\textsc{Has21\_Hi}} & 0.1727 & 0.0808 & 79.17 & 5.6 \\
        \textbf{\textsc{Ous19\_Es}} & 0.1852 & 0.0869 & 69.09 & 6.02 \\
        \textbf{\textsc{San20\_It}} & 0.4004 & 0.019 & 40.73 & 1.32 \\
        \bottomrule
        \multicolumn{5}{l}{} \\
        \multicolumn{5}{l}{Model: \textsc{X+DEn}} \\
        \toprule
        \textbf{Test Set} &  \bm{$\beta_0$} &  \bm{$\beta_1$} & \bm{$R^2$} & \bm{$2N$}\\
        \midrule
        \textbf{\textsc{Bas19\_Es}} & 0.5543 & 0.0363 & 87.75 & 2.52 \\
        \textbf{\textsc{For19\_Pt}} & 0.5756 & 0.0205 & 63.66 & 1.42 \\
        \textbf{\textsc{Has21\_Hi}} & 0.4691 & 0.0163 & 43.91 & 1.13 \\
        \textbf{\textsc{Ous19\_Es}} & 0.4320 & 0.0391 & 64.99 & 2.71 \\
        \textbf{\textsc{San20\_It}} & 0.5724 & 0.0272 & 69.42 & 1.89 \\
        \bottomrule
        \multicolumn{5}{l}{} \\
        \multicolumn{5}{l}{Model: \textsc{X+FEn}} \\
        \toprule
        \textbf{Test Set} &  \bm{$\beta_0$} &  \bm{$\beta_1$} & \bm{$R^2$} & \bm{$2N$}\\
        \midrule
        \textbf{\textsc{Bas19\_Es}} & 0.4347 & 0.0489 & 93.95 & 3.39 \\
        \textbf{\textsc{For19\_Pt}} & 0.6167 & 0.0148 & 60.96 & 1.02 \\
        \textbf{\textsc{Has21\_Hi}} & 0.5684 & 0.0192 & 45.73 & 1.33 \\
        \textbf{\textsc{Ous19\_Es}} & 0.6256 & 0.0191 & 63.71 & 1.32 \\
        \textbf{\textsc{San20\_It}} & 0.5227 & 0.008 & 7.81 & 0.56 \\
        \bottomrule
        \multicolumn{5}{l}{} \\
        \multicolumn{5}{l}{Model: \textsc{X+KEn}} \\
        \toprule
        \textbf{Test Set} &  \bm{$\beta_0$} &  \bm{$\beta_1$} & \bm{$R^2$} & \bm{$2N$}\\
        \midrule
        \textbf{\textsc{Bas19\_Es}} & 0.5011 & 0.0381 & 93.59 & 2.64 \\
        \textbf{\textsc{For19\_Pt}} & 0.5797 & 0.0182 & 72.81 & 1.26 \\
        \textbf{\textsc{Has21\_Hi}} & 0.5545 & 0.0195 & 40.66 & 1.35 \\
        \textbf{\textsc{Ous19\_Es}} & 0.5712 & 0.026 & 82.84 & 1.8 \\
        \textbf{\textsc{San20\_It}} & 0.4444 & 0.0198 & 42.5 & 1.37 \\
        \bottomrule
    \end{tabular}
    }
\caption{OLS results for all models in our experiments, expanding on Table~\ref{tab: ols} from the main body.
All regression coefficients $\beta_0$ and  $\beta_1$ are significant at $p<0.001$.} \label{tab: ols_full}
\end{table}

\setlength{\tabcolsep}{4.5pt}
\begin{table*}[h]
    \centering
    \resizebox{0.9\textwidth}{!}{ 
    \begin{tabular}{l|cccccccccccc}
            \multicolumn{1}{c}{} & \multicolumn{12}{c}{\textbf{\textsc{Bas19\_Es}}}\\
        \toprule
            \multicolumn{1}{r|}{\textbf{N}} & \textbf{10} & \textbf{20} & \textbf{30} & \textbf{40} & \textbf{50} & \textbf{100} & \textbf{200} & \textbf{300} & \textbf{400} & \textbf{500} & \textbf{1000} & \textbf{2,000}\\
        \midrule
            \textbf{M} & 0.46 & 0.50 & 0.56 & 0.57 & 0.59 & 0.66 & 0.72 & \textbf{0.77} & \textbf{0.79} & \textbf{0.80} & \textbf{0.83} & \textbf{0.84} \\
            \textbf{X} & 0.41 & 0.48 & 0.49 & 0.46 & 0.55 & 0.60 & 0.70 & 0.74 & 0.76 & 0.77 & 0.79 & 0.81 \\
            \textbf{\textsc{X+DEn}} & \textbf{0.63} & \textbf{0.66} & \textbf{0.68} & \textbf{0.69} & \textbf{0.70} & \textbf{0.72} & \textbf{0.75} & 0.76 & 0.78 & 0.78 & 0.81 & 0.82 \\
            \textbf{\textsc{X+FEn}} & 0.56 & 0.59 & 0.60 & 0.61 & 0.62 & 0.65 & 0.68 & 0.71 & 0.74 & 0.75 & 0.78 & 0.80 \\
            \textbf{\textsc{X+KEn}} & 0.61 & 0.61 & 0.63 & 0.63 & 0.63 & 0.67 & 0.70 & 0.72 & 0.74 & 0.74 & 0.77 & 0.79 \\
        \bottomrule
            \multicolumn{13}{c}{} \\
            \multicolumn{1}{c}{} & \multicolumn{12}{c}{\textbf{\textsc{For19\_Pt}}}\\
        \toprule
            \multicolumn{1}{r|}{\textbf{N}} & \textbf{10} & \textbf{20} & \textbf{30} & \textbf{40} & \textbf{50} & \textbf{100} & \textbf{200} & \textbf{300} & \textbf{400} & \textbf{500} & \textbf{1000} & \textbf{2,000}\\
        \midrule
            \textbf{M} & 0.45 & 0.46 & 0.44 & 0.46 & 0.50 & 0.59 & 0.67 & 0.68 & 0.70 & 0.69 & 0.71 & \textbf{0.73} \\
            \textbf{X} & 0.40 & 0.42 & 0.41 & 0.42 & 0.45 & 0.51 & 0.67 & 0.68 & 0.70 & 0.70 & 0.71 & \textbf{0.73} \\
            \textbf{\textsc{X+DEn}} & 0.62 & 0.63 & 0.64 & 0.64 & 0.65 & 0.68 & 0.70 & \textbf{0.70} & 0.70 & 0.70 & \textbf{0.72} & 0.72 \\
            \textbf{\textsc{X+FEn}} & \textbf{0.66} & \textbf{0.66} & \textbf{0.66} & \textbf{0.67} & \textbf{0.66} & \textbf{0.69} & \textbf{0.71} & \textbf{0.70} & \textbf{0.71} & \textbf{0.71} & \textbf{0.72} & \textbf{0.73} \\
            \textbf{\textsc{X+KEn}} & 0.62 & 0.65 & 0.64 & 0.64 & 0.64 & 0.66 & 0.69 & 0.69 & 0.69 & 0.70 & 0.71 & 0.71 \\
        \bottomrule
            \multicolumn{13}{c}{} \\
            \multicolumn{1}{c}{} & \multicolumn{12}{c}{\textbf{\textsc{Has21\_Hi}}}\\
        \toprule
            \multicolumn{1}{r|}{\textbf{N}} & \textbf{10} & \textbf{20} & \textbf{30} & \textbf{40} & \textbf{50} & \textbf{100} & \textbf{200} & \textbf{300} & \textbf{400} & \textbf{500} & \textbf{1000} & \textbf{2,000}\\
        \midrule
            \textbf{M} & 0.47 & 0.47 & 0.47 & 0.47 & 0.47 & 0.50 & 0.55 & 0.55 & 0.56 & \textbf{0.57} & 0.58 & \textbf{0.60} \\
            \textbf{X} & 0.47 & 0.47 & 0.47 & 0.47 & 0.47 & 0.47 & 0.47 & 0.47 & 0.50 & 0.54 & 0.56 & 0.56 \\
            \textbf{\textsc{X+DEn}} & 0.52 & 0.52 & 0.53 & 0.53 & 0.52 & 0.53 & \textbf{0.56} & 0.56 & 0.56 & \textbf{0.57} & \textbf{0.59} & \textbf{0.60} \\
            \textbf{\textsc{X+FEn}} & \textbf{0.55} & \textbf{0.55} & \textbf{0.56} & \textbf{0.55} & \textbf{0.55} & \textbf{0.54} & \textbf{0.56} & \textbf{0.57} & \textbf{0.57} & \textbf{0.57} & 0.58 & 0.59 \\
            \textbf{\textsc{X+KEn}} & 0.50 & 0.52 & 0.51 & 0.51 & 0.49 & 0.52 & \textbf{0.56} & 0.56 & 0.58 & 0.55 & 0.58 & \textbf{0.60} \\
        \bottomrule
            \multicolumn{13}{c}{} \\
            \multicolumn{1}{c}{} & \multicolumn{12}{c}{\textbf{\textsc{Ous19\_Ar}}}\\
        \toprule
            \multicolumn{1}{r|}{\textbf{N}} & \textbf{10} & \textbf{20} & \textbf{30} & \textbf{40} & \textbf{50} & \textbf{100} & \textbf{200} & \textbf{300} & \textbf{400} & \textbf{500} & \textbf{1000} & \textbf{2,000}\\
        \midrule
            \textbf{M} & 0.48 & 0.48 & 0.46 & 0.51 & 0.53 & 0.53 & 0.66 & 0.64 & \textbf{0.70} & \textbf{0.71} & \textbf{0.72} & \textbf{0.72} \\
            \textbf{X} & 0.42 & 0.43 & 0.43 & 0.43 & 0.43 & 0.49 & \textbf{0.69} & \textbf{0.69} & 0.69 & 0.69 & 0.71 & 0.70 \\
            \textbf{\textsc{X+DEn}} & 0.54 & 0.52 & 0.53 & 0.58 & 0.59 & 0.63 & 0.66 & 0.68 & 0.68 & 0.69 & 0.68 & 0.70 \\
            \textbf{\textsc{X+FEn}} & \textbf{0.61} & \textbf{0.61} & \textbf{0.63} & \textbf{0.65} & \textbf{0.64} & \textbf{0.67} & 0.68 & \textbf{0.69} & 0.69 & 0.68 & 0.70 & 0.70 \\
            \textbf{\textsc{X+KEn}} & 0.60 & 0.60 & 0.62 & 0.62 & 0.62 & 0.65 & 0.67 & 0.67 & 0.67 & 0.68 & 0.69 & 0.69 \\
        \bottomrule
            \multicolumn{13}{c}{} \\
            \multicolumn{1}{c}{} & \multicolumn{12}{c}{\textbf{\textsc{San20\_It}}}\\
        \toprule
            \multicolumn{1}{r|}{\textbf{N}} & \textbf{10} & \textbf{20} & \textbf{30} & \textbf{40} & \textbf{50} & \textbf{100} & \textbf{200} & \textbf{300} & \textbf{400} & \textbf{500} & \textbf{1000} & \textbf{2,000}\\
        \midrule
            \textbf{M} & 0.44 & 0.40 & 0.44 & 0.50 & 0.56 & 0.69 & \textbf{0.73} & \textbf{0.75} & \textbf{0.75} & \textbf{0.76} & \textbf{0.78} & \textbf{0.79} \\
            \textbf{X} & 0.45 & 0.40 & 0.35 & 0.49 & 0.56 & 0.63 & 0.70 & 0.73 & 0.74 & 0.74 & 0.77 & 0.78 \\
            \textbf{\textsc{X+DEn}} & 0.64 & 0.63 & 0.67 & 0.67 & 0.68 & 0.71 & \textbf{0.73} & 0.73 & 0.74 & 0.74 & 0.76 & 0.77 \\
            \textbf{\textsc{X+FEn}} & \textbf{0.68} & \textbf{0.66} & \textbf{0.69} & \textbf{0.69} & \textbf{0.71} & \textbf{0.72} & \textbf{0.73} & 0.74 & \textbf{0.75} & 0.74 & 0.76 & 0.76 \\
            \textbf{\textsc{X+KEn}} & 0.64 & 0.64 & 0.66 & 0.65 & 0.67 & 0.68 & 0.71 & 0.72 & 0.73 & 0.74 & 0.75 & 0.76 \\
        \bottomrule
    \end{tabular}
    }
\caption{Macro F1 on respective held-out test sets for models fine-tuned on N target-language entries, averaged across 10 random seeds for each N. Best performance for a given N in \textbf{bold}.}
\label{tab: macroF1_testsets_fullpage}
\end{table*}

\begin{figure*}[h]
\centering
\includegraphics[height=0.9\textheight]{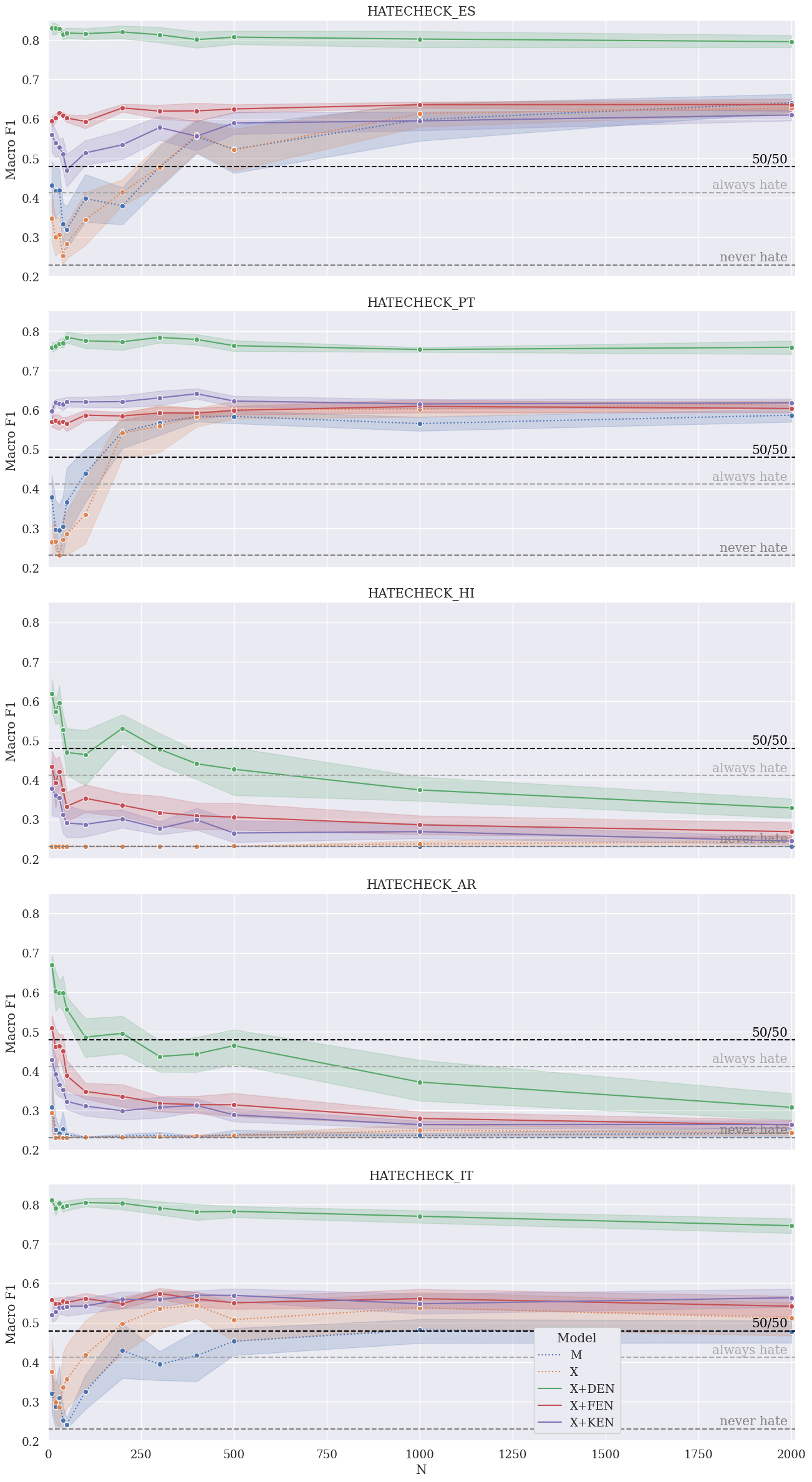}
\caption{
Macro F1 on target-language tests from MHC across models trained on up to N = 2,000 target-language entries.
Model notation as described in \S\ref{subsec: models}.
Confidence intervals and model baselines as described in \S\ref{subsec: held-out testsets}.
}
\label{fig: macroF1_hatecheck_fullpage}
\end{figure*}

\setlength{\tabcolsep}{4.5pt}
\begin{table*}[h]
    \centering
    \resizebox{0.9\textwidth}{!}{ 
    \begin{tabular}{l|cccccccccccc}
            \multicolumn{1}{c}{} & \multicolumn{12}{c}{\textbf{\textsc{HateCheck\_Es}}}\\
        \toprule
            \multicolumn{1}{r|}{\textbf{N}} & \textbf{10} & \textbf{20} & \textbf{30} & \textbf{40} & \textbf{50} & \textbf{100} & \textbf{200} & \textbf{300} & \textbf{400} & \textbf{500} & \textbf{1000} & \textbf{2,000}\\
        \midrule
            M & 0.43 & 0.42 & 0.42 & 0.33 & 0.32 & 0.40 & 0.38 & 0.48 & 0.56 & 0.52 & 0.60 & 0.64 \\
            X & 0.35 & 0.30 & 0.31 & 0.25 & 0.28 & 0.34 & 0.41 & 0.48 & 0.56 & 0.52 & 0.61 & 0.63 \\
            X+DEN & \textbf{0.83} & \textbf{0.83} & \textbf{0.83} & \textbf{0.82} & \textbf{0.82} & \textbf{0.82} & \textbf{0.82} & \textbf{0.81} & \textbf{0.80} & \textbf{0.81} & \textbf{0.80} & \textbf{0.80} \\
            X+FEN & 0.59 & 0.60 & 0.61 & 0.61 & 0.60 & 0.59 & 0.63 & 0.62 & 0.62 & 0.63 & 0.64 & 0.64 \\
            X+KEN & 0.56 & 0.54 & 0.53 & 0.51 & 0.47 & 0.51 & 0.53 & 0.58 & 0.56 & 0.59 & 0.59 & 0.61 \\
        \bottomrule
            \multicolumn{13}{c}{} \\
            \multicolumn{1}{c}{} & \multicolumn{12}{c}{\textbf{\textsc{HateCheck\_Pt}}}\\
        \toprule
            \multicolumn{1}{r|}{\textbf{N}} & \textbf{10} & \textbf{20} & \textbf{30} & \textbf{40} & \textbf{50} & \textbf{100} & \textbf{200} & \textbf{300} & \textbf{400} & \textbf{500} & \textbf{1000} & \textbf{2,000}\\
        \midrule
            M & 0.38 & 0.30 & 0.30 & 0.31 & 0.37 & 0.44 & 0.54 & 0.57 & 0.58 & 0.58 & 0.57 & 0.59 \\
            X & 0.27 & 0.27 & 0.23 & 0.27 & 0.28 & 0.33 & 0.54 & 0.56 & 0.58 & 0.60 & 0.60 & 0.62 \\
            X+DEN & \textbf{0.76} & \textbf{0.76} & \textbf{0.77} & \textbf{0.77} & \textbf{0.78} & \textbf{0.78} & \textbf{0.77} & \textbf{0.78} & \textbf{0.78} & \textbf{0.76} & \textbf{0.75} & \textbf{0.76} \\
            X+FEN & 0.57 & 0.57 & 0.57 & 0.57 & 0.57 & 0.59 & 0.58 & 0.59 & 0.59 & 0.60 & 0.61 & 0.60 \\
            X+KEN & 0.60 & 0.62 & 0.62 & 0.62 & 0.62 & 0.62 & 0.62 & 0.63 & 0.64 & 0.62 & 0.62 & 0.62 \\
        \bottomrule
            \multicolumn{13}{c}{} \\
            \multicolumn{1}{c}{} & \multicolumn{12}{c}{\textbf{\textsc{HateCheck\_Hi}}}\\
        \toprule
            \multicolumn{1}{r|}{\textbf{N}} & \textbf{10} & \textbf{20} & \textbf{30} & \textbf{40} & \textbf{50} & \textbf{100} & \textbf{200} & \textbf{300} & \textbf{400} & \textbf{500} & \textbf{1000} & \textbf{2,000}\\
        \midrule
            M & 0.23 & 0.23 & 0.23 & 0.23 & 0.23 & 0.23 & 0.23 & 0.23 & 0.23 & 0.23 & 0.23 & 0.23 \\
            X & 0.23 & 0.23 & 0.23 & 0.23 & 0.23 & 0.23 & 0.23 & 0.23 & 0.23 & 0.23 & 0.24 & 0.24 \\
            X+DEN & \textbf{0.62} & \textbf{0.57} & \textbf{0.60} & \textbf{0.53} & \textbf{0.47} & \textbf{0.46} & \textbf{0.53} & \textbf{0.48} & \textbf{0.44} & \textbf{0.43} & \textbf{0.37} & \textbf{0.33} \\
            X+FEN & 0.43 & 0.39 & 0.42 & 0.37 & 0.33 & 0.35 & 0.34 & 0.32 & 0.31 & 0.31 & 0.29 & 0.27 \\
            X+KEN & 0.38 & 0.36 & 0.35 & 0.31 & 0.29 & 0.29 & 0.30 & 0.28 & 0.30 & 0.27 & 0.27 & 0.24 \\
        \bottomrule
            \multicolumn{13}{c}{} \\
            \multicolumn{1}{c}{} & \multicolumn{12}{c}{\textbf{\textsc{HateCheck\_Ar}}}\\
        \toprule
            \multicolumn{1}{r|}{\textbf{N}} & \textbf{10} & \textbf{20} & \textbf{30} & \textbf{40} & \textbf{50} & \textbf{100} & \textbf{200} & \textbf{300} & \textbf{400} & \textbf{500} & \textbf{1000} & \textbf{2,000}\\
        \midrule
            M & 0.31 & 0.25 & 0.24 & 0.25 & 0.24 & 0.23 & 0.23 & 0.24 & 0.23 & 0.24 & 0.24 & 0.24 \\
            X & 0.29 & 0.23 & 0.23 & 0.23 & 0.23 & 0.23 & 0.23 & 0.23 & 0.24 & 0.24 & 0.25 & 0.24 \\
            X+DEN & \textbf{0.67} & \textbf{0.60} & \textbf{0.60} & \textbf{0.60} & \textbf{0.56} & \textbf{0.49} & \textbf{0.50} & \textbf{0.44} & \textbf{0.44} & \textbf{0.46} & \textbf{0.37} & \textbf{0.31} \\
            X+FEN & 0.51 & 0.46 & 0.46 & 0.45 & 0.39 & 0.35 & 0.34 & 0.32 & 0.31 & 0.31 & 0.28 & 0.26 \\
            X+KEN & 0.43 & 0.39 & 0.37 & 0.35 & 0.32 & 0.31 & 0.30 & 0.31 & 0.31 & 0.29 & 0.26 & 0.26 \\
        \bottomrule
            \multicolumn{13}{c}{} \\
            \multicolumn{1}{c}{} & \multicolumn{12}{c}{\textbf{\textsc{HateCheck\_It}}}\\
        \toprule
            \multicolumn{1}{r|}{\textbf{N}} & \textbf{10} & \textbf{20} & \textbf{30} & \textbf{40} & \textbf{50} & \textbf{100} & \textbf{200} & \textbf{300} & \textbf{400} & \textbf{500} & \textbf{1000} & \textbf{2,000}\\
        \midrule
            M & 0.32 & 0.29 & 0.31 & 0.25 & 0.24 & 0.33 & 0.43 & 0.39 & 0.42 & 0.45 & 0.48 & 0.48 \\
            X & 0.38 & 0.30 & 0.29 & 0.34 & 0.36 & 0.42 & 0.50 & 0.54 & 0.54 & 0.51 & 0.54 & 0.51 \\
            X+DEN & \textbf{0.81} & \textbf{0.79} & \textbf{0.80} & \textbf{0.79} & \textbf{0.80} & \textbf{0.80} & \textbf{0.80} & \textbf{0.79} & \textbf{0.78} & \textbf{0.78} & \textbf{0.77} & \textbf{0.75} \\
            X+FEN & 0.56 & 0.55 & 0.55 & 0.55 & 0.55 & 0.56 & 0.55 & 0.57 & 0.56 & 0.55 & 0.56 & 0.54 \\
            X+KEN & 0.52 & 0.53 & 0.54 & 0.54 & 0.54 & 0.54 & 0.56 & 0.56 & 0.57 & 0.57 & 0.55 & 0.56 \\
        \bottomrule
    \end{tabular}
    }
\caption{Macro F1 on target-language HateCheck for models fine-tuned on N target-language entries, averaged across 10 random seeds for each N. Best performance for a given N in \textbf{bold}.}
\label{tab: macroF1_hatecheck_fullpage}
\end{table*}

\end{document}